\documentclass[sigconf, screen]{acmart}
\usepackage{booktabs}
\usepackage[normalem]{ulem}
\useunder{\uline}{\ul}{}
\usepackage{multirow} 
\usepackage{subcaption}
\usepackage{graphicx}

\usepackage{algorithm}
\usepackage{algpseudocode}

\usepackage{mathtools}
\usepackage{float}

\AtBeginDocument{%
  }

\setcopyright{acmlicensed}
\copyrightyear{2018}
\acmYear{2018}
\acmDOI{XXXXXXX.XXXXXXX}

\acmConference[Conference acronym 'XX]{Make sure to enter the correct
  conference title from your rights confirmation emai}{June 03--05,
  2018}{Woodstock, NY}
\acmISBN{978-1-4503-XXXX-X/18/06}

\begin{document}

\title{MetaWild: A Multimodal Dataset for Animal Re-Identification with Environmental Metadata}

\author{Yuzhuo Li}
\authornote{Equal contribution}
\affiliation{%
  \institution{University of Auckland}
  \city{Auckland}
  \country{New Zealand}
}
\email{yil708@aucklanduni.ac.nz}

\author{Di Zhao}
\authornotemark[1]
\authornote{Corresponding author}
\affiliation{%
  \institution{University of Auckland}
  \city{Auckland}
  \country{New Zealand}
}
\email{dzha866@aucklanduni.ac.nz}

\author{Tingrui Qiao}
\affiliation{%
  \institution{University of Auckland}
  \city{Auckland}
  \country{New Zealand}
}
\email{tqia361@aucklanduni.ac.nz}

\author{Yihao Wu}
\affiliation{%
  \institution{University of Auckland}
  \city{Auckland}
  \country{New Zealand}
}
\email{ywu840@aucklanduni.ac.nz}

\author{Bo Pang}
\affiliation{%
  \institution{University of Auckland}
  \city{Auckland}
  \country{New Zealand}
}
\email{bpan882@aucklanduni.ac.nz}

\author{Yun Sing Koh}
\affiliation{%
  \institution{University of Auckland}
  \city{Auckland}
  \country{New Zealand}
}
\email{y.koh@auckland.ac.nz}

\begin{abstract}
Identifying individual animals within large wildlife populations is essential for effective wildlife monitoring and conservation efforts.
Recent advancements in computer vision have shown promise in animal re-identification (Animal ReID) by leveraging data from camera traps.
However, existing Animal ReID datasets rely exclusively on visual data, overlooking environmental metadata that ecologists have identified as highly correlated with animal behavior and identity, such as temperature and circadian rhythms.
Moreover, the emergence of multimodal models capable of jointly processing visual and textual data presents new opportunities for Animal ReID, but existing benchmarks fail to leverage these models' text-processing capabilities, limiting their full potential. 
To address these limitations, we propose the MetaWild dataset, a multimodal Animal ReID benchmark pairing images with environmental metadata extracted from embedded camera trap overlays and scene contexts. 
MetaWild contains 20,890 images spanning six representative species, providing a robust resource for systematically evaluating multimodal approaches in Animal ReID.
Additionally, to facilitate the use of metadata in existing ReID methods, we propose the Meta-Feature Adapter (MFA), a lightweight module that can be incorporated into existing vision-language model (VLM)-based Animal ReID methods, allowing ReID models to leverage both environmental metadata and visual information to improve ReID performance.
Experiments on MetaWild show that combining baseline ReID models with MFA to incorporate metadata consistently improves performance compared to using visual information alone, validating the effectiveness of incorporating metadata in re-identification. 
We hope that our proposed dataset can inspire further exploration of multimodal approaches for Animal ReID.
Our dataset and supplementary materials are available at \url{https://jim-lyz1024.github.io/MetaWild/}.

\end{abstract}



\maketitle

\section{Introduction}
Animal re-identification (Animal ReID) aims to recognize and match individual animals across different images, enabling researchers to track specific individuals over time and across locations~\cite{schneider2019past}. 
It is essential for studying various aspects of wildlife, such as population monitoring, movement ecology, behavioral studies, and wildlife management~\cite{papafitsoros2021social,schofield2022more}.
Compared to monitoring traditional methods such as physical tagging, scarring, or branding, recent advances in computer vision have clear advantages as they offer a non-invasive approach to monitor endangered species without causing stress or behavior changes~\cite{beery2023wild,xu2024advanced}.

\begin{figure}[!t]
    \centering
    \includegraphics[width=0.90\columnwidth]{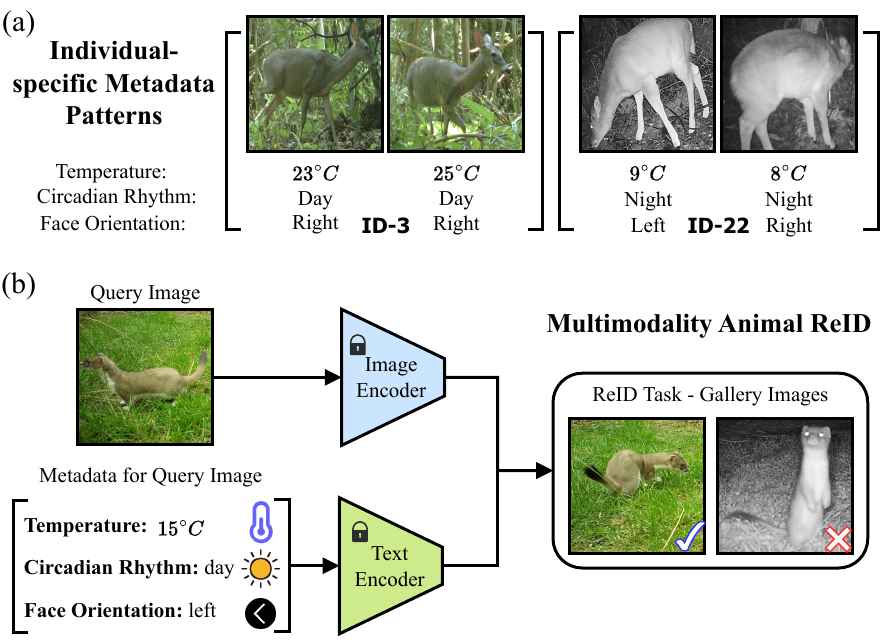}
    \caption{Overview of multimodal Animal ReID framework and the role of metadata. (a) Example images of two deer individuals showing distinct preferences across environmental conditions. (b) The model integrates visual features with textual environmental metadata. }
    \label{fig:three_methods}
\end{figure}

In Animal ReID, the dataset is crucially important to fairly and precisely evaluate the performance of the ReID methods.
Most existing datasets are constructed from images captured by surveillance cameras, handheld devices, or camera traps, and are presented purely in visual form~\cite{adam2024seaturtleid2022,li2020atrw,gao2021towards}.
This visual-only design introduces two limitations.
First, it omits environmental metadata (\textit{e.g.}, temperature and circadian rhythms), which ecologists have shown to be highly correlated with animal behavior and appearance~\cite{leliveld2022dairy}.
Figure~\ref{fig:three_methods}(a) shows that different individuals tend to appear under distinct environmental conditions, showing the potential of metadata to provide identity-discriminative cues.
Without such information, existing datasets cannot support evaluating the impact of incorporating environmental metadata on ReID performance.
Second, with the rapid advancement of multimodal models capable of jointly processing images and text~\cite{radford2021learning,jia2021scaling}, current image-only datasets restrict these models to operating solely through their image encoders, leaving the text encoder underexplored and limiting the full potential of multimodal representations in Animal ReID.

To address these limitations, we present MetaWild, a dataset designed to enable systematic evaluation of metadata integration and multimodal learning in Animal ReID.
MetaWild is constructed from the publicly available New Zealand Trail Camera (NZ-TrailCams) dataset~\cite{nztrailcams}, a wildlife image collection captured using camera traps deployed in diverse natural habitats across New Zealand. 
To ensure broad ecological coverage and metadata diversity, we select six representative species, including invasive and native animals.
We carefully curated images that are visually clear and contain reliable metadata overlays, with each species sufficiently represented (minimum 2,433 images per species) and containing 20,890 images in total. 
In addition to visual data, MetaWild includes curated environmental metadata extracted from embedded camera trap overlays and scene contexts, including \textit{temperature}, \textit{circadian rhythms}, and \textit{face orientation}, chosen based on their availability from overlays and influence on animal appearance.
This metadata enables the construction of a multimodal Animal ReID framework, where visual and textual cues are jointly exploited, as illustrated in Figure~\ref{fig:three_methods}(b).

While the MetaWild dataset enables the investigation of multimodal Animal ReID, most existing Animal ReID methods are generally designed to operate solely on visual inputs and cannot directly incorporate the textual metadata. 
To address this gap, we propose the Meta-Feature Adapter (MFA), a lightweight module that can be incorporated into existing vision-language model (VLM)-based Animal ReID methods~\cite{jiao2024toward}, allowing the integration of environmental metadata into visual representations.
Specifically, MFA translates structured metadata into natural language descriptions, embeds them using the text encoder from VLM, and injects the resulting metadata-aware text embeddings into the visual feature space through a gated cross-attention mechanism, which uses a learnable gating function to suppress noisy or redundant metadata~\cite{zhang2024visual}.

Our \textbf{contributions} are summarized as follows.
Firstly, we create and release MetaWild, a new dataset that pairs visual animal images with curated environmental metadata to evaluate the impact of incorporating environmental metadata on ReID performance and facilitates the development of multimodal ReID methods based on VLMs.
Secondly, we propose the Meta-Feature Adapter (MFA), a simple yet effective module that enables integration of metadata into existing VLM-based Animal ReID models, allowing performance evaluation with MetaWild without modifying model architecture.
Finally, extensive experiments on the MetaWild dataset and ReID models combined with MFA show that incorporating environmental metadata alongside visual data within a multimodal learning framework can significantly improve Animal ReID performance.

\section{Related Work}
\noindent \textbf{Animal ReID} aims to recognize and match individual animals across images, supporting wildlife monitoring and conservation. 
Existing methods can be categorized into three types:
(1) Global Feature Learning: These methods treat the entire image as input, directly extracting global features for ReID~\cite{he2023animal}. 
While adaptable across species, they often rely on techniques originally developed for person ReID.
(2) Species-Specific Feature Extraction: Leveraging distinctive local patterns (\textit{e.g.}, whale tails~\cite{cheeseman2022advanced}, elephant ears~\cite{weideman2020extracting}), these methods enhance identification by focusing on species-specific features. 
However, they are sensitive to occlusions and viewpoint variations, which hinder their generalizability across species.
(3) Auxiliary Information Integration: Methods such as pose key point estimation~\cite{li2020atrw} and head pose recognition~\cite{zhang2021yakreid} incorporate additional visual cues to refine feature extraction. 
Despite their success, these methods remain constrained to image-based data.

\noindent \textbf{Animal ReID Datasets.}
Numerous datasets have been developed to advance Animal ReID, ranging from controlled environment datasets that focus on animals in farms and laboratories~\cite{kern2024towards,gao2021towards,wahltinez2024open}, to in-the-wild collections that capture animals in natural habitats, introducing real-world challenges such as occlusions and viewpoint variations~\cite{li2020atrw,adam2024seaturtleid2022,adam2025exploiting}.
More recent efforts, WildlifeDatasets~\cite{vcermak2024wildlifedatasets}, provide a unified library and standardized tools for loading, preprocessing, and evaluating ReID datasets across diverse species and scenarios.
While these datasets have advanced visual-based ReID, they omit environmental metadata, which has been shown to influence animal behavior and appearance, limiting the evaluation of its impact on ReID performance.

\noindent \textbf{Vision-Language Model} 
(VLM) aims to build a cohesive alignment between images and languages to learn a shared embedding space encompassing both modalities~\cite{radford2021learning,jia2021scaling,alayrac2022flamingo}.
CLIP (Contrastive Language-Image Pre-Training)~\cite{radford2021learning} is the most widely used VLM and has shown impressive zero-shot capabilities and robust generalization in various vision-language tasks~\cite{chen2024large,chen2024ssat,zhao2024symmetric,li2023clip}.
While some recent works have adapted VLMs to ReID tasks~\cite{li2023clip,jiao2024toward}, they primarily leverage the image encoder and use the text encoder only for static, category-level descriptions.
In this paper, we leverage environmental metadata as a semantically rich textual information source to more fully utilize the text encoder for Animal ReID.

\section{The MetaWild Dataset}
\subsection{Dataset Composition}
The MetaWild dataset is designed to facilitate the evaluation of multimodal learning approaches in Animal ReID by pairing visual data with contextual environmental metadata. 
MetaWild is constructed from the NZ-TrailCams dataset~\cite{nztrailcams}, a large-scale wildlife image collection publicly available through the Labeled Information Library of Alexandria: Biology and Conservation (LILA)~\cite{lila}.
To ensure the applicability and diversity of metadata integration across various animal species, MetaWild comprises 20,890 images spanning six representative species: Deer, Hare, Penguin, Pūkeko, Stoat, and Wallaby.
Each image is paired with environmental metadata extracted from embedded camera trap overlays.
The MetaWild dataset follows the same licensing terms as the original NZ-TrailCams dataset, under the CDLA-Permissive-1.0~\cite{cdla}.
\begin{figure}[ht]
    \centering
    \includegraphics[width=1.0\linewidth]{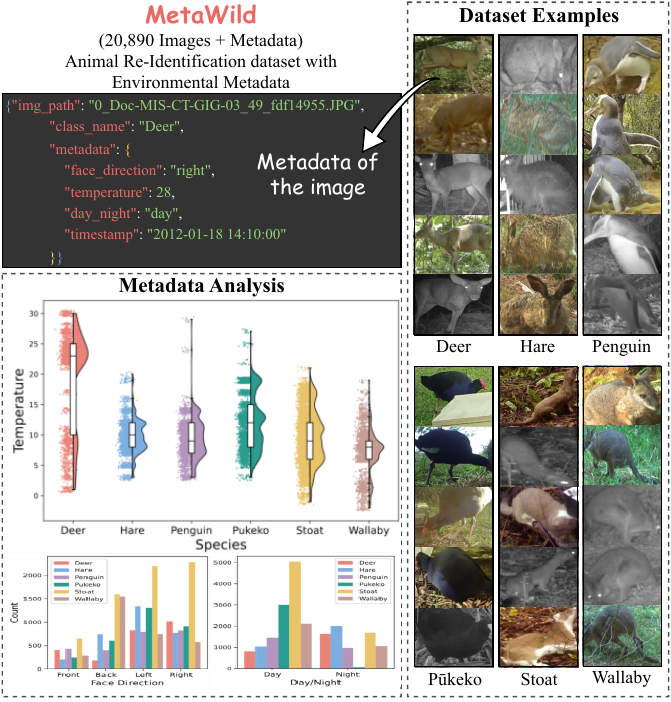}
    \caption{Examples of the MetaWild dataset.}
    \label{fig:dataset_illustration}
\end{figure}

Figure~\ref{fig:dataset_illustration} presents examples of the species and environmental metadata included in the MetaWild dataset. 
The selection of species reflects critical conservation priorities, including predators (Stoat), pests (Wallaby, Hare), endangered native species (Yellow-eyed Penguin), and native animals (Deer and Pūkeko)~\cite{docpests,docnative}.
Stoats compete with native birdlife for food and habitat, also eat the eggs and young, and attack the adults, posing a significant threat to native wildlife. 
The Yellow-eyed Penguin, classified as endangered with a rapidly declining population, exemplifies a vulnerable native species in urgent need of protection. 
Wallabies and hares are regarded as agricultural pests, causing substantial damage to native vegetation and ecosystems. 
Deer and Pūkeko, both native to New Zealand, present unique visual and behavioral patterns that introduce diversity and realism to further enrich the dataset's applicability.
The dataset was organized according to standard ReID protocols, comprising 60\% for training, 25\% for the gallery, and 15\% for the query sets~\cite{li2018richly}. 
Detailed statistics for each species are provided in Table~\ref{tab:benchmark_detail}.
\begin{table}[ht]
\caption{Details of Benchmark Datasets.}
\label{tab:benchmark_detail}
\centering
\fontsize{9pt}{9pt}\selectfont
\begin{tabular}{@{}l|rrrrrr|rr@{}}
\toprule
 & \multicolumn{2}{c}{Train} & \multicolumn{2}{c}{Gallery} & \multicolumn{2}{c|}{Query} & \multicolumn{2}{c}{Total} \\ \midrule
Datasets & Imgs & IDs & Imgs & IDs & Imgs & IDs & Imgs & IDs \\ \midrule
Deer & 1,631 & 21 & 586 & 17 & 216 & 17 & 2,433 & 38 \\
Hare & 1,820 & 31 & 926 & 29 & 306 & 29 & 3,052 & 60 \\
Penguin & 1,431 & 34 & 725 & 43 & 296 & 43 & 2,452 & 77 \\
Pūkeko & 1,854 & 11 & 800 & 19 & 411 & 19 & 3,065 & 30 \\
Stoat & 4,067 & 151 & 1,649 & 102 & 1,017 & 102 & 6,733 & 253 \\
Wallaby & 1,888 & 25 & 964 & 22 & 303 & 22 & 3,155 & 47 \\ \bottomrule
\end{tabular}
\end{table}

\subsection{Dataset Construction}
To ensure high data quality for Animal ReID, we focused on selecting visually clear images, providing accurate identity annotations, and curating reliable environmental metadata. 
The construction process involved image selection, identity annotation, metadata extraction, and image preprocessing.
To ensure annotation reliability, both identity labels and metadata were independently verified by at least three annotators.
The detailed procedure is described below.

\noindent \textbf{Image Selection.}
The images from the NZ-TrailCams~\cite{nztrailcams} were gathered by a variety of conservation projects using different camera trap brands and configurations.
We first conducted a filtering process to remove low-quality images in which the target animals in images appeared as unrecognizable blurry blobs due to motion blur or poor lighting.
We further selected a representative subset of images for each species to ensure sufficient intra-species variation across individuals and conditions (\textit{e.g.}, viewpoints, lighting, and environments) and maintain balanced distribution across environmental metadata.
Following this process, we selected a high-quality, balanced subset of approximately 3,000 images per species to construct the MetaWild dataset.

\noindent \textbf{Identity Annotation.}
Annotating individual animal identities is challenging and essential for ReID tasks, particularly when ground truth labels are unavailable.
We employed a combination of temporal analysis and visual verification to assign identities within each species.
For temporal analysis, we used the camera trap images' time-stamped nature to track animals across sequential frames.
Animals captured within narrow time windows (\textit{e.g.}, a few seconds) at the same camera location were likely to belong to the same individual.
To complement this, manual visual inspection was conducted to confirm or correct identity groupings based on distinct physical characteristics, such as markings, size, and shape.
Species with prominent features, such as the unique plumage patterns of the yellow-eyed penguin, allowed for more straightforward identification.
On average, each identity comprises approximately 120 images, ensuring sufficient data for training and evaluation.

\begin{figure}[ht]
    \centering
    \begin{subfigure}[b]{0.11\textwidth}
        \centering
        \includegraphics[width=\textwidth]{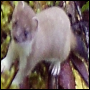}
        \caption{Front}
        \label{fig:front}
    \end{subfigure}
    \begin{subfigure}[b]{0.11\textwidth}
        \centering
        \includegraphics[width=\textwidth]{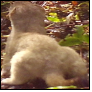}
        \caption{Back}
        \label{fig:back}
    \end{subfigure}
    \begin{subfigure}[b]{0.11\textwidth}
        \centering
        \includegraphics[width=\textwidth]{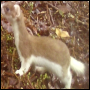}
        \caption{Left}
        \label{fig:left}
    \end{subfigure}
    \begin{subfigure}[b]{0.11\textwidth}
        \centering
        \includegraphics[width=\textwidth]{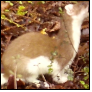}
        \caption{Right}
        \label{fig:right}
    \end{subfigure}
    \caption{Example images showing different face orientations in the MetaWild dataset.}
    \label{fig:face_orientations}
\end{figure}
\noindent \textbf{Metadata Extraction.}
A key innovation of our dataset is the integration of environmental metadata extracted from camera trap images.
We focus on three metadata features: \textit{temperature}, \textit{circadian rhythms}, and \textit{face orientation}, chosen for their direct influence on the animal's appearance and behavior~\cite{cade2021tools,mcvey2023invited}.
Temperature, read from embedded camera trap overlays, affects activity patterns and may alter visible characteristics, such as fur density or coloration~\cite{cossins2012temperature}.
Face orientation, manually annotated from image content, provides geometric context essential for feature matching in the Animal ReID tasks (Figure~\ref{fig:face_orientations})~\cite{adam2025exploiting}.
Circadian rhythm is defined as a binary variable (day/night) based on the image's timestamp and lighting conditions in the scene, which affect image quality and the visibility of distinguishing features.
To maintain metadata quality and relevance, we selectively excluded features that were either redundant or inaccessible.
For other metadata available in the overlays, such as atmospheric pressure, it was excluded due to minimal variation and limited relevance, as they are largely determined by temperature and altitude.
Furthermore, while other metafeatures such as rainfall, humidity, terrain, and proximity to water sources could potentially improve Animal ReID by capturing environmental variation across locations, these cannot be extracted in our setting because the NZ-TrailCams dataset provides only \textit{CameraIDs} without associated geographic coordinates.
All metadata is standardized and stored in accompanying JSON files for each image. 

\noindent \textbf{Image Preprocessing.}
To focus on the animal subject and minimize background distraction, we cropped the original images to retain only the regions containing the target animal.  
We implemented this by training a YOLO-based object detection model to generate bounding boxes around animals in the images and using the bounding boxes to crop each image around the detected animal automatically.
All bounding boxes were manually verified to correct for false positives, occlusions, or overlapping instances.
Furthermore, each cropped image was renamed using a structured format, $id\_camera$-$id\_count$ (\textit{e.g.}, $11\_CT$-$GIG$-$03\_27$, where $11$ denotes the individual identity, $CT$-$GIG$-$03$ represents the camera ID, and $27$ indicates the 28th image for identity 11).
This systematic naming convention facilitates efficient data management and traceability.

\begin{figure*}[ht]
    \centering    \includegraphics[width=0.85\textwidth]{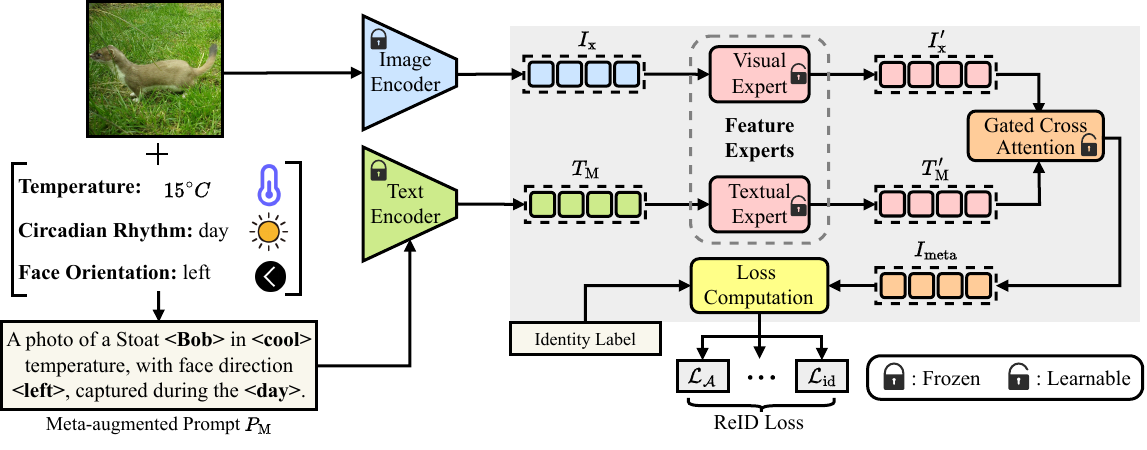}
    \caption{Overview of the proposed Meta-Feature Adapter (MFA) module}
    \label{fig:MFA_framework}
\end{figure*}
\section{Methodology}
To investigate the impact of environmental metadata introduced in the MetaWild dataset without redesigning ReID models from scratch, we propose the Meta-Feature Adapter (MFA), a lightweight module that enables metadata integration into existing VLM-based Animal ReID frameworks.
The overview of the proposed MFA module is illustrated in Figure~\ref{fig:MFA_framework}. 
The MFA consists of two key components:  
(1) \textbf{Feature Experts}, employed adapters~\cite{gao2024clip} as experts in both the image and text branches to ensure that both modalities capture metadata-aware representations.
(2) \textbf{Gated Cross-Attention}, which fuses visual and metadata features by weighting relevant information from each modality.

\subsection{Feature Experts}
To enable effective fusion between visual features and environmental metadata, we incorporate feature experts in both the text and image branches.
These modules are designed to refine the original embeddings produced by the VLMs' encoder and produce representations that are better aligned for cross-modal interaction. 
While VLMs offer strong pretrained encoders, their raw outputs may contain irrelevant or redundant information and are not tailored for integration with metadata. 
Directly combining such features can hinder downstream ReID performance. 

In the text branch, to incorporate metadata into the VLM's text encoder, we first convert metadata into natural language descriptions using a fixed prompt template: ``A photo of a \{species\} \{individual id\} in \{freezing, cold, chilly, cool, warm, hot\} temperature, with face direction \{front, back, left, right\}, captured during the \{day, night\}."
This template mimics natural language captions used during VLM pretraining, which helps maximize compatibility with the text encoder and enables more effective metadata fusion into the shared embedding space.
Here, we convert numerical metadata Temperature into categorical descriptors [freezing, cold, chilly, cool, warm, hot]~\cite{gagge1967comfort} as numerical metadata often lacks intuitive contextual meaning.
This metadata-augmented prompt $P_{\text{M}}$ is then encoded using the pretrained text encoder $\mathcal{T}(\cdot)$ to obtain the corresponding text embedding $T_{\text{M}} = \mathcal{T}(P_\text{M})$.
To further adapt this embedding for the ReID task, we employ a Textual Metadata Expert (TME), represented as $E_T$, to refine $T_{\text{M}}$ into a metadata-aware embedding $T'_{\text{M}}$ by $T_\text{M}' = E_\text{T}(T_\text{M})$.

In the image branch, we similarly introduce a Visual Feature Expert (VFE), $E_I$, to adapt the raw visual embedding $I_\text{x}$ from the image encoder by making it more responsive to metadata.
Similar to the TME, the VFE is an adapter to transform the image embeddings $I_\text{x}$ into metadata-aware visual representations $I_\text{x}'$ by $I_\text{x}' = E_I(I_\text{x})$.
Both experts $E_T$ and $E_I$ are implemented as lightweight MLP-based adapters~\cite{gao2024clip} with residual connections.
The feature experts $E_T$ and $E_I$ are trained end-to-end using the same ReID loss functions adopted by the baseline models. 
These losses typically include the identity classification loss $\mathcal{L}_{id}$, which encourages separability across individual identities, and the triplet loss $\mathcal{L}_{tri}$, which enforces relative distance constraints in the embedding space by pulling positive pairs closer while pushing negative pairs farther apart. 
Depending on the baseline configuration, additional losses such as contrastive loss may also be used~\cite{wang2021understanding}. 
By allowing gradients from the ReID objective to propagate through the feature experts, $E_T$ and $E_I$ are guided to learn task-relevant transformations that directly contribute to improving ReID performance.

\subsection{Gated Cross-Attention}
We utilize a cross-attention mechanism to integrate metadata-aware text embeddings $T'_{\text{M}}$ with visual features $I'_{\text{x}}$.
Unlike conventional feature concatenation or fusion methods, cross-attention enables context-aware integration, allowing different image feature components to attend to relevant metadata cues selectively.
Given the image embeddings $I_\text{x}' \in \mathbb{R}^{N \times d}$ produced by VFE and metadata-aware text embeddings $T_\text{M}' \in \mathbb{R}^{M \times d}$, we compute the Query ($Q$), Key ($K$), and Value ($V$) matrices with Equation \eqref{eq:query_key_value}.
\begin{equation}
\label{eq:query_key_value}
Q = I_\text{x}' W_Q, \quad K = T_\text{M}' W_K, \quad V = T_\text{M}' W_V
\end{equation}
Here, $W_Q$, $W_K$, and $W_V$ are learnable projection matrices.
We then compute cross-attention weights $A$~\cite{shi2022dense}.

While cross-attention enables alignment between modalities, it treats all metadata contributions equally, ignoring the fact that the relevance of metadata can vary across images.
To allow the model to incorporate metadata into the visual representation selectively, we employ a gating mechanism that can adjust the contribution of metadata based on its relevance to each image.
We compute a gating value $\gamma \in [0, 1]$ to determine the contribution of metadata, as defined in Equation~\eqref{eq:gated_attention}.
\begin{equation}
\label{eq:gated_attention}
\gamma = \text{Gate}(I_\text{x}', T_\text{M}') = \sigma(\text{MLP}([I_\text{x}'; T_\text{M}']))
\end{equation}
Here, $[I_\text{x}'; T_\text{M}']$ denotes the concatenation of image and metadata embeddings.
The MLP represents a multi-layer perception with layer normalization, while $\sigma$ is the activation function (sigmoid), ensuring that $\gamma$ is constrained between $[0, 1]$.
The final meta-augmented image embedding, $I_{\text{meta}}$, is then computed by Equation~\eqref{eq:gated_imagefeature}.
\begin{equation}
\label{eq:gated_imagefeature}
I_{\text{meta}} = (\text{Gate}(I_\text{x}', T_\text{M}') \odot (AV)) + I_\text{x}' = \gamma AV + I_\text{x}'
\end{equation}
Here, $\odot$ represents element-wise multiplication, and $A$ is the attention weight matrix.
The loss of the cross-attention module is:
\begin{equation}
    \label{eq:loss_attention}
    \mathcal{L}_{\mathcal{A}}^i = -\log \frac{\exp\left( s\left( T_{\text{M}}'^{i}, I_{\text{meta}}^{i} \right) / \tau \right)}{\sum_{j=1}^{B} \exp\left( s\left( T_{\text{M}}'^i, I_{\text{meta}}^{j} \right) / \tau \right)}
\end{equation}
Here, $B$ represents the batch size, $( T_{M}'^i, I_{\text{meta}}^i)$ denotes the $i\text{-th}$ matched pair of metadata-aware text embeddings and meta-augmented image embeddings, and $\tau$ is the temperature parameter.

\section{Experiments}
We evaluate existing Animal ReID models under visual-only and visual+metadata settings on the MetaWild dataset to assess the impact of environmental metadata on ReID performance.
We evaluate our approach using two protocols: 
(1) \textbf{Intra-species ReID}~\cite{varghese2023fine}, where training and testing are performed on different individuals within the same species; and 
(2) \textbf{Inter-species ReID}~\cite{heiling2016using},  where we adopt a leave-one-domain-out evaluation strategy~\cite{yu2024rethinking}, in which the model is trained on five species and evaluated on the remaining unseen species. 
This setup reflects real-world scenarios where collecting labeled data for every species is impractical and offers deeper insights into the model's ability to extract species-agnostic identity features~\cite{jiao2024toward}.

\subsection{Experimental Results}
\noindent \textbf{Intra-species Re-Identification Results.}
Table~\ref{tab:evaluation_results_intra} compares the intra-species ReID performance of models under visual-only and visual+metadata settings across six species.
Across all six species, incorporating environmental metadata consistently improves ReID performance. 
For example, CLIP-ReID achieves mAP gains of 5.5\% on Penguin, 4.9\% on Wallaby, and 4.2\% on Deer after metadata integration. 
ReID-AW shows similar improvements, with notable gains of 6.5\% on Penguin, 5.1\% on Wallaby, and 4.9\% on Deer.
These consistent improvements demonstrate the effectiveness of incorporating metadata for improving Animal ReID performance under the intra-species protocol.

\begin{table*}[ht]
\caption{Intra-species re-identification performance on the MetaWild dataset across six species, we report mAP and CMC-1 accuracy (\%) with 95\% confidence intervals. CLIP-ZS shows zero variance due to its deterministic zero-shot inference nature.}
\label{tab:evaluation_results_intra}
\centering
\fontsize{9pt}{9pt}\selectfont
\setlength{\tabcolsep}{1.0mm}
\begin{tabular}{@{}l|cc|cc|cc|cc|cc|cc@{}}
\toprule
\multirow{2}{*}{Methods} & \multicolumn{2}{c|}{Deer} & \multicolumn{2}{c|}{Hare} & \multicolumn{2}{c|}{Penguin} & \multicolumn{2}{c|}{Pūkeko} & \multicolumn{2}{c|}{Stoat} & \multicolumn{2}{c}{Wallaby} \\ 
 & mAP & CMC-1 & mAP & CMC-1 & mAP & CMC-1 & mAP & CMC-1 & mAP & CMC-1 & mAP & CMC-1 \\ \midrule
CLIP-ZS~\cite{radford2021learning} & 50.0±.0 & 92.1±.0 & 33.8±.0 & 81.1±.0 & 34.0±.0 & 57.8±.0 & 33.0±.0 & 69.6±.0 & 30.1±.0 & 72.7±.0 & 46.2±.0 & 85.5±.0 \\ \midrule
CLIP-FT & 63.2±.1 & 95.4±.4 & 56.7±.2 & 92.5±.4 & 44.0±.3 & 64.9±.2 & 56.8±.1 & 80.3±.5 & 68.6±.1 & 92.3±.3 & 55.5±.1 & 92.4±.4 \\
CLIP-FT+MFA & \textbf{66.7±.2} & \textbf{95.7±.3} & \textbf{58.4±.3} & 92.6±.2 & \textbf{46.0±.2} & 64.9±.3 & \textbf{58.2±.2} & \textbf{81.6±.3} & \textbf{69.8±.2} & 91.8±.2 & \textbf{56.8±.4} & 90.7±.3 \\ \midrule
CLIP-ReID~\cite{li2023clip} & 65.2±.4 & 95.8±.3 & 60.0±.6 & 95.1±.3 & 44.8±.4 & 67.9±.4 & 57.6±.2 & 82.0±.1 & 67.5±.1 & 91.5±.3 & 56.9±.4 & 88.8±.2 \\
CLIP-ReID+MFA & \textbf{69.4±.2} & \textbf{98.1±.1} & \textbf{63.2±.1} & 95.4±.1 & \textbf{50.3±.4} & \textbf{68.6±.2} & \textbf{59.8±.1} & \textbf{83.7±.2} & \textbf{71.5±.2} & \textbf{92.0±.1} & \textbf{61.8±.2} & \textbf{92.1±.1} \\ \midrule
ReID-AW~\cite{jiao2024toward} & \multicolumn{1}{l}{67.5±.3} & 96.0±.2 & 63.3±.4 & 95.6±.3 & 48.8±.5 & 69.4±.3 & 58.5±.3 & 82.0±.4 & 69.5±.3 & 93.5±.5 & 58.4±.3 & 91.8±.2 \\
ReID-AW+MFA & \textbf{72.4±.2} & \textbf{97.0±.2} & \textbf{66.2±.3} & \textbf{96.8±.2} & \textbf{55.3±.4} & \textbf{70.8±.4} & \textbf{61.8±.2} & \textbf{86.7±.3} & \textbf{74.1±.4} & \textbf{95.0±.2} & \textbf{63.5±.1} & \textbf{92.7±.2} \\ \bottomrule
\end{tabular}
\end{table*}

\noindent \textbf{Inter-species Re-Identification Results.}
Table~\ref{tab:evaluation_results_cross} summarizes the results of leave-one-domain-out evaluations conducted on all six species in the MetaWild dataset. 
In each evaluation round, one species is held out as the target domain for testing, while the remaining five are used for training.
Each column in the table corresponds to the evaluation scenario where the respective species is treated as the unseen target domain.
Across all settings, incorporating metadata consistently improves the performance of all baseline methods. 
For example, CLIP-ReID with metadata (CLIP-ReID+MFA) achieves mAP gains of 3.9\% on Deer, 3.3\% on Pūkeko, and 1.8\% on Penguin compared to its visual-only counterpart. 
ReID-AW, the strongest baseline, also benefits from metadata integration, with mAP improvements of 3.4\% on Penguin, 3.2\% on Deer, and 2.6\% on Hare. 
These results show that environmental metadata offers complementary cues that help models generalize across species boundaries, enhancing their ability to extract transferable identity representations in low-supervision settings.
\begin{table*}[ht]
\caption{Leave-one-domain-out inter-species ReID performance on the MetaWild dataset, we report mAP and CMC-1 accuracy (\%) with 95\% confidence intervals for each target species.}
\label{tab:evaluation_results_cross}
\centering
\fontsize{9pt}{9pt}\selectfont
\setlength{\tabcolsep}{1.0mm}
\begin{tabular}{@{}l|cc|cc|cc|cc|cc|cc@{}}
\toprule
\multirow{2}{*}{Methods} & \multicolumn{2}{c|}{Deer} & \multicolumn{2}{c|}{Hare} & \multicolumn{2}{c|}{Penguin} & \multicolumn{2}{c|}{Pūkeko} & \multicolumn{2}{c|}{Stoat} & \multicolumn{2}{c}{Wallaby} \\ 
 & mAP & CMC-1 & mAP & CMC-1 & mAP & CMC-1 & mAP & CMC-1 & mAP & CMC-1 & mAP & CMC-1 \\ \midrule
CLIP-ZS~\cite{radford2021learning} & 40.3±.0 & 80.5±.0 & 25.1±.0 & 75.1±.0 & 27.5±.0 & 50.4±.0 & 24.6±.0 & 64.5±.0 & 23.0±.0 & 62.3±.0 & 40.6±.0 & 80.3±.0 \\ \midrule
CLIP-FT & 55.1±.2 & 83.0±.4 & 41.5±.2 & 81.4±.2 & 38.7±.3 & 59.7±.2 & 41.3±.2 & 76.9±.2 & 45.4±.3 & 79.3±.2 & 49.0±.2 & 72.2±.1 \\
CLIP-FT+MFA & \textbf{56.7±.2} & \textbf{86.4±.4} & \textbf{43.9±.4} & 81.6±.2 & \textbf{40.8±.1} & \textbf{63.5±.2} & \textbf{42.3±.3} & \textbf{76.9±.4} & \textbf{46.2±.4} & 79.6±.2 & \textbf{50.0±.3} & \textbf{72.8±.3} \\ \midrule
CLIP-ReID~\cite{li2023clip} & 56.3±.3 & 84.4±.3 & 43.5±.1 & 86.3±.3 & 39.5±.2 & 60.3±.4 & 43.8±.4 & 77.9±.1 & 45.8±.3 & 79.3±.2 & 50.1±.4 & 82.5±.2 \\
CLIP-ReID+MFA & \textbf{60.2±.4} & \textbf{89.2±.3} & \textbf{44.1±.1} & \textbf{88.6±.2} & \textbf{41.3±.2} & \textbf{64.1±.4} & \textbf{45.1±.3} & \textbf{78.3±.2} & \textbf{47.9±.1} & \textbf{80.3±.2} & \textbf{52.3±.2} & \textbf{82.8±.2} \\ \midrule
ReID-AW~\cite{jiao2024toward} & 59.3±.3 & 89.0±.2 & 47.6±.2 & 90.2±.4 & 40.8±.5 & 63.9±.3 & 50.4±.4 & 80.3±.1 & 53.3±.3 & 83.9±.1 & 51.7±.3 & 84.2±.2 \\
ReID-AW+MFA & \textbf{62.5±.4} & \textbf{92.4±.4} & \textbf{50.2±.3} & \textbf{90.8±.2} & \textbf{44.2±.4} & \textbf{64.6±.3} & \textbf{53.6±.3} & \textbf{83.6±.1} & \textbf{56.1±.1} & \textbf{84.6±.1} & \textbf{53.1±.2} & \textbf{86.0±.3} \\ \bottomrule
\end{tabular}
\end{table*}

\subsection{Qualitative Analysis}
\begin{figure}[ht]
    \centering
    \includegraphics[width=1.0\linewidth]{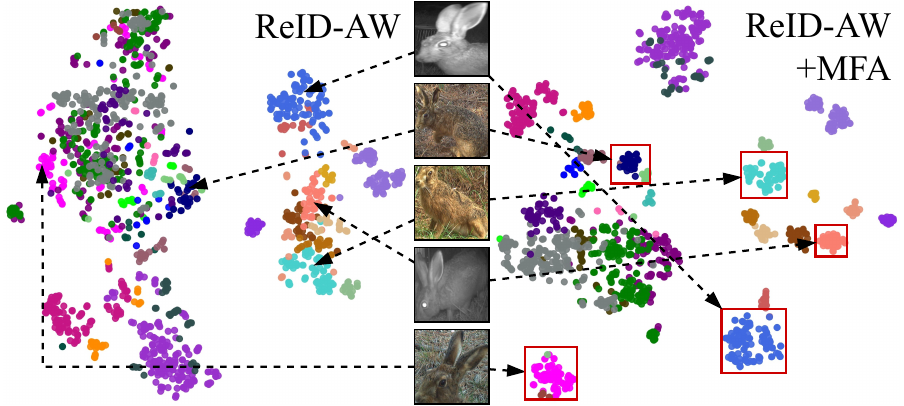}
    \caption{t-SNE visualization of learned embeddings on Hare. Each color indicates a different individual.}
    \label{fig:TSNE}
\end{figure}

\begin{figure}[ht]
\centering
\includegraphics[width=0.75\linewidth]{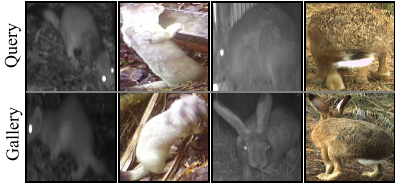}
\caption{Challenging ReID cases where incorporating metadata resolves failures under visual-only settings. Each column shows a query image (top) and its corresponding correct match from the gallery (bottom) for the same individual.}
\label{fig:challenging_cases}
\end{figure}

\noindent \textbf{T-SNE Visualization}.
To better understand the impact of metadata integration on ReID performance, we visualize the embedding distributions using t-SNE~\cite{van2008visualizing} on the Hare dataset for both ReID-AW and ReID-AW+MFA models.
Figure~\ref{fig:TSNE} shows the distribution of embeddings where each color represents a different individual, with representative images displayed for selected challenging cases.
ReID-AW+MFA shows more compact and well-separated clusters for the same individuals than the baseline, suggesting that metadata integration helps the model learn more discriminative embeddings.
\noindent
\textbf{Visualization of Challenging Scenarios}.
To further illustrate the impact of metadata, we present qualitative examples in Figure~\ref{fig:challenging_cases}.
Each column displays a challenging case in which the visual-only model fails to match the query image (top) with its correct counterpart in the gallery (bottom), while correct identification is achieved when metadata is incorporated.
In the night scenarios (first and third columns), the circadian rhythm metadata guides the model to focus on features that remain distinctive under low-light conditions, such as body shapes, rather than being misled by illumination variations. 
Similarly, in cases with consistent face orientations (second and fourth columns), the orientation metadata helps the model identify orientation-specific characteristics. 
For example, focusing on tail patterns when viewing from the back or facial marks when viewing from the front.
These examples show how metadata provides complementary information that helps improve ReID performance.

\section{Conclusion}
In this paper, we present MetaWild, a multimodal dataset for Animal ReID that pairs visual data with textual environmental metadata. 
MetaWild is specifically designed to evaluate the impact of environmental metadata on Animal ReID performance.
To support this investigation without requiring changes to existing model architectures, we further propose the Meta-Feature Adapter (MFA), a lightweight module that enables the integration of metadata into VLM-based Animal ReID methods. 
Extensive experiments demonstrate that incorporating metadata alongside visual information consistently improves ReID accuracy, confirming the value of contextual environmental metadata. 
We hope this work can inspire broader exploration of environmental metadata and multimodal approaches in wildlife ReID and beyond.


\bibliographystyle{ACM-Reference-Format}
\bibliography{main}


\begin{thebibliography}{43}


\ifx \showCODEN    \undefined \def \showCODEN     #1{\unskip}     \fi
\ifx \showISBNx    \undefined \def \showISBNx     #1{\unskip}     \fi
\ifx \showISBNxiii \undefined \def \showISBNxiii  #1{\unskip}     \fi
\ifx \showISSN     \undefined \def \showISSN      #1{\unskip}     \fi
\ifx \showLCCN     \undefined \def \showLCCN      #1{\unskip}     \fi
\ifx \shownote     \undefined \def \shownote      #1{#1}          \fi
\ifx \showarticletitle \undefined \def \showarticletitle #1{#1}   \fi
\ifx \showURL      \undefined \def \showURL       {\relax}        \fi
\providecommand\bibfield[2]{#2}
\providecommand\bibinfo[2]{#2}
\providecommand\natexlab[1]{#1}
\providecommand\showeprint[2][]{arXiv:#2}

\bibitem[lil(2024)]%
        {lila}
 \bibinfo{year}{Accessed: 2024}\natexlab{}.
\newblock \bibinfo{booktitle}{\emph{Labeled Information Library of Alexandria: Biology and Conservation (LILA)}}.
\newblock
\urldef\tempurl%
\url{https://lila.science/}
\showURL{%
\tempurl}


\bibitem[nzt(2024)]%
        {nztrailcams}
 \bibinfo{year}{Accessed: 2024}\natexlab{}.
\newblock \bibinfo{booktitle}{\emph{Trail Camera Images of New Zealand Animals}}.
\newblock
\urldef\tempurl%
\url{https://lila.science/datasets/nz-trailcams}
\showURL{%
\tempurl}


\bibitem[cdl(2025)]%
        {cdla}
 \bibinfo{year}{Accessed: 2025}\natexlab{}.
\newblock \bibinfo{booktitle}{\emph{Community Data License Agreement – Permissive – Version 1.0.}}
\newblock
\urldef\tempurl%
\url{https://cdla.dev/permissive-1-0/}
\showURL{%
\tempurl}


\bibitem[doc(2025a)]%
        {docnative}
 \bibinfo{year}{Accessed: 2025}\natexlab{a}.
\newblock \bibinfo{booktitle}{\emph{New Zealand's native animals.}}
\newblock
\urldef\tempurl%
\url{https://www.doc.govt.nz/nature/native-animals/}
\showURL{%
\tempurl}


\bibitem[doc(2025b)]%
        {docpests}
 \bibinfo{year}{Accessed: 2025}\natexlab{b}.
\newblock \bibinfo{booktitle}{\emph{New Zealand's unique biodiversity is at risk from pests, weeds and other threats.}}
\newblock
\urldef\tempurl%
\url{https://www.doc.govt.nz/nature/pests-and-threats/}
\showURL{%
\tempurl}


\bibitem[Adam et~al\mbox{.}(2024)]%
        {adam2024seaturtleid2022}
\bibfield{author}{\bibinfo{person}{Luk{\'a}{\v{s}} Adam}, \bibinfo{person}{Vojt{\v{e}}ch {\v{C}}erm{\'a}k}, \bibinfo{person}{Kostas Papafitsoros}, {and} \bibinfo{person}{Lukas Picek}.} \bibinfo{year}{2024}\natexlab{}.
\newblock \showarticletitle{SeaTurtleID2022: A long-span dataset for reliable sea turtle re-identification}. In \bibinfo{booktitle}{\emph{Proceedings of the IEEE/CVF Winter Conference on Applications of Computer Vision}}. \bibinfo{pages}{7146--7156}.
\newblock


\bibitem[Adam et~al\mbox{.}(2025)]%
        {adam2025exploiting}
\bibfield{author}{\bibinfo{person}{Luk{\'a}{\v{s}} Adam}, \bibinfo{person}{Kostas Papafitsoros}, \bibinfo{person}{Claire Jean}, \bibinfo{person}{ALan~F Rees}, {and} \bibinfo{person}{Vojt{\v{e}}ch {\v{C}}erm{\'a}k}.} \bibinfo{year}{2025}\natexlab{}.
\newblock \showarticletitle{Exploiting facial side similarities to improve AI-driven sea turtle photo-identification systems}.
\newblock \bibinfo{journal}{\emph{Ecological Informatics}} (\bibinfo{year}{2025}), \bibinfo{pages}{103158}.
\newblock


\bibitem[Alayrac et~al\mbox{.}(2022)]%
        {alayrac2022flamingo}
\bibfield{author}{\bibinfo{person}{Jean-Baptiste Alayrac}, \bibinfo{person}{Jeff Donahue}, \bibinfo{person}{Pauline Luc}, \bibinfo{person}{Antoine Miech}, \bibinfo{person}{Iain Barr}, \bibinfo{person}{Yana Hasson}, \bibinfo{person}{Karel Lenc}, \bibinfo{person}{Arthur Mensch}, \bibinfo{person}{Katherine Millican}, \bibinfo{person}{Malcolm Reynolds}, {et~al\mbox{.}}} \bibinfo{year}{2022}\natexlab{}.
\newblock \showarticletitle{Flamingo: a visual language model for few-shot learning}.
\newblock \bibinfo{journal}{\emph{Advances in Neural Information Processing Systems}}  \bibinfo{volume}{35} (\bibinfo{year}{2022}), \bibinfo{pages}{23716--23736}.
\newblock


\bibitem[Beery(2023)]%
        {beery2023wild}
\bibfield{author}{\bibinfo{person}{Sara~Meghan Beery}.} \bibinfo{year}{2023}\natexlab{}.
\newblock \bibinfo{booktitle}{\emph{Where the Wild Things Are: Computer Vision for Global-Scale Biodiversity Monitoring}}.
\newblock \bibinfo{publisher}{California Institute of Technology}.
\newblock


\bibitem[Cade et~al\mbox{.}(2021)]%
        {cade2021tools}
\bibfield{author}{\bibinfo{person}{David~E Cade}, \bibinfo{person}{William~T Gough}, \bibinfo{person}{Max~F Czapanskiy}, \bibinfo{person}{James~A Fahlbusch}, \bibinfo{person}{Shirel~R Kahane-Rapport}, \bibinfo{person}{Jacob~MJ Linsky}, \bibinfo{person}{Ross~C Nichols}, \bibinfo{person}{William~K Oestreich}, \bibinfo{person}{Danuta~M Wisniewska}, \bibinfo{person}{Ari~S Friedlaender}, {et~al\mbox{.}}} \bibinfo{year}{2021}\natexlab{}.
\newblock \showarticletitle{Tools for integrating inertial sensor data with video bio-loggers, including estimation of animal orientation, motion, and position}.
\newblock \bibinfo{journal}{\emph{Animal Biotelemetry}} \bibinfo{volume}{9}, \bibinfo{number}{1} (\bibinfo{year}{2021}), \bibinfo{pages}{34}.
\newblock


\bibitem[{\v{C}}erm{\'a}k et~al\mbox{.}(2024)]%
        {vcermak2024wildlifedatasets}
\bibfield{author}{\bibinfo{person}{Vojt{\v{e}}ch {\v{C}}erm{\'a}k}, \bibinfo{person}{Lukas Picek}, \bibinfo{person}{Luk{\'a}{\v{s}} Adam}, {and} \bibinfo{person}{Kostas Papafitsoros}.} \bibinfo{year}{2024}\natexlab{}.
\newblock \showarticletitle{WildlifeDatasets: An open-source toolkit for animal re-identification}. In \bibinfo{booktitle}{\emph{Proceedings of the IEEE/CVF Winter Conference on Applications of Computer Vision}}. \bibinfo{pages}{5953--5963}.
\newblock


\bibitem[Cheeseman et~al\mbox{.}(2022)]%
        {cheeseman2022advanced}
\bibfield{author}{\bibinfo{person}{Ted Cheeseman}, \bibinfo{person}{Ken Southerland}, \bibinfo{person}{Jinmo Park}, \bibinfo{person}{Marilia Olio}, \bibinfo{person}{Kiirsten Flynn}, \bibinfo{person}{John Calambokidis}, \bibinfo{person}{Lindsey Jones}, \bibinfo{person}{Claire Garrigue}, \bibinfo{person}{Astrid Frisch~Jord{\'a}n}, \bibinfo{person}{Addison Howard}, {et~al\mbox{.}}} \bibinfo{year}{2022}\natexlab{}.
\newblock \showarticletitle{Advanced image recognition: a fully automated, high-accuracy photo-identification matching system for humpback whales}.
\newblock \bibinfo{journal}{\emph{Mammalian Biology}} \bibinfo{volume}{102}, \bibinfo{number}{3} (\bibinfo{year}{2022}), \bibinfo{pages}{915--929}.
\newblock


\bibitem[Chen et~al\mbox{.}(2024a)]%
        {chen2024ssat}
\bibfield{author}{\bibinfo{person}{Bowen Chen}, \bibinfo{person}{Yun~Sing Koh}, {and} \bibinfo{person}{Gillian Dobbie}.} \bibinfo{year}{2024}\natexlab{a}.
\newblock \showarticletitle{SSAT-Adapter: Enhancing Vision-Language Model Few-shot Learning with Auxiliary Tasks}. In \bibinfo{booktitle}{\emph{Proceedings of the 32nd ACM International Conference on Multimedia}}. \bibinfo{pages}{1004--1013}.
\newblock


\bibitem[Chen et~al\mbox{.}(2024b)]%
        {chen2024large}
\bibfield{author}{\bibinfo{person}{Liangyu Chen}, \bibinfo{person}{Bo Li}, \bibinfo{person}{Sheng Shen}, \bibinfo{person}{Jingkang Yang}, \bibinfo{person}{Chunyuan Li}, \bibinfo{person}{Kurt Keutzer}, \bibinfo{person}{Trevor Darrell}, {and} \bibinfo{person}{Ziwei Liu}.} \bibinfo{year}{2024}\natexlab{b}.
\newblock \showarticletitle{Large language models are visual reasoning coordinators}.
\newblock \bibinfo{journal}{\emph{Advances in Neural Information Processing Systems}}  \bibinfo{volume}{36} (\bibinfo{year}{2024}).
\newblock


\bibitem[Cossins(2012)]%
        {cossins2012temperature}
\bibfield{author}{\bibinfo{person}{Andrew Cossins}.} \bibinfo{year}{2012}\natexlab{}.
\newblock \bibinfo{booktitle}{\emph{Temperature biology of animals}}.
\newblock \bibinfo{publisher}{Springer Science \& Business Media}.
\newblock


\bibitem[Gagge et~al\mbox{.}(1967)]%
        {gagge1967comfort}
\bibfield{author}{\bibinfo{person}{A~Pharo Gagge}, \bibinfo{person}{Jan~AJ Stolwijk}, {and} \bibinfo{person}{James~Daniel Hardy}.} \bibinfo{year}{1967}\natexlab{}.
\newblock \showarticletitle{Comfort and thermal sensations and associated physiological responses at various ambient temperatures}.
\newblock \bibinfo{journal}{\emph{Environmental Research}} \bibinfo{volume}{1}, \bibinfo{number}{1} (\bibinfo{year}{1967}), \bibinfo{pages}{1--20}.
\newblock


\bibitem[Gao et~al\mbox{.}(2021)]%
        {gao2021towards}
\bibfield{author}{\bibinfo{person}{Jing Gao}, \bibinfo{person}{Tilo Burghardt}, \bibinfo{person}{William Andrew}, \bibinfo{person}{Andrew~W Dowsey}, {and} \bibinfo{person}{Neill~W Campbell}.} \bibinfo{year}{2021}\natexlab{}.
\newblock \showarticletitle{Towards self-supervision for video identification of individual holstein-friesian cattle: The Cows2021 dataset}.
\newblock \bibinfo{journal}{\emph{arXiv preprint arXiv:2105.01938}} (\bibinfo{year}{2021}).
\newblock


\bibitem[Gao et~al\mbox{.}(2024)]%
        {gao2024clip}
\bibfield{author}{\bibinfo{person}{Peng Gao}, \bibinfo{person}{Shijie Geng}, \bibinfo{person}{Renrui Zhang}, \bibinfo{person}{Teli Ma}, \bibinfo{person}{Rongyao Fang}, \bibinfo{person}{Yongfeng Zhang}, \bibinfo{person}{Hongsheng Li}, {and} \bibinfo{person}{Yu Qiao}.} \bibinfo{year}{2024}\natexlab{}.
\newblock \showarticletitle{Clip-adapter: Better vision-language models with feature adapters}.
\newblock \bibinfo{journal}{\emph{International Journal of Computer Vision}} \bibinfo{volume}{132}, \bibinfo{number}{2} (\bibinfo{year}{2024}), \bibinfo{pages}{581--595}.
\newblock


\bibitem[He et~al\mbox{.}(2023)]%
        {he2023animal}
\bibfield{author}{\bibinfo{person}{Zhimin He}, \bibinfo{person}{Jiangbo Qian}, \bibinfo{person}{Diqun Yan}, \bibinfo{person}{Chong Wang}, {and} \bibinfo{person}{Yu Xin}.} \bibinfo{year}{2023}\natexlab{}.
\newblock \showarticletitle{Animal re-identification algorithm for posture diversity}. In \bibinfo{booktitle}{\emph{ICASSP 2023-2023 IEEE International Conference on Acoustics, Speech and Signal Processing}}. IEEE, \bibinfo{pages}{1--5}.
\newblock


\bibitem[Heiling et~al\mbox{.}(2016)]%
        {heiling2016using}
\bibfield{author}{\bibinfo{person}{Sven Heiling}, \bibinfo{person}{Santosh Khanal}, \bibinfo{person}{Aiko Barsch}, \bibinfo{person}{Gabriela Zurek}, \bibinfo{person}{Ian~T Baldwin}, {and} \bibinfo{person}{Emmanuel Gaquerel}.} \bibinfo{year}{2016}\natexlab{}.
\newblock \showarticletitle{Using the knowns to discover the unknowns: MS-based dereplication uncovers structural diversity in 17-hydroxygeranyllinalool diterpene glycoside production in the Solanaceae}.
\newblock \bibinfo{journal}{\emph{The Plant Journal}} \bibinfo{volume}{85}, \bibinfo{number}{4} (\bibinfo{year}{2016}), \bibinfo{pages}{561--577}.
\newblock


\bibitem[Jia et~al\mbox{.}(2021)]%
        {jia2021scaling}
\bibfield{author}{\bibinfo{person}{Chao Jia}, \bibinfo{person}{Yinfei Yang}, \bibinfo{person}{Ye Xia}, \bibinfo{person}{Yi-Ting Chen}, \bibinfo{person}{Zarana Parekh}, \bibinfo{person}{Hieu Pham}, \bibinfo{person}{Quoc Le}, \bibinfo{person}{Yun-Hsuan Sung}, \bibinfo{person}{Zhen Li}, {and} \bibinfo{person}{Tom Duerig}.} \bibinfo{year}{2021}\natexlab{}.
\newblock \showarticletitle{Scaling up visual and vision-language representation learning with noisy text supervision}. In \bibinfo{booktitle}{\emph{International Conference on Machine Learning}}. PMLR, \bibinfo{pages}{4904--4916}.
\newblock


\bibitem[Jiao et~al\mbox{.}(2024)]%
        {jiao2024toward}
\bibfield{author}{\bibinfo{person}{Bingliang Jiao}, \bibinfo{person}{Lingqiao Liu}, \bibinfo{person}{Liying Gao}, \bibinfo{person}{Ruiqi Wu}, \bibinfo{person}{Guosheng Lin}, \bibinfo{person}{Peng Wang}, {and} \bibinfo{person}{Yanning Zhang}.} \bibinfo{year}{2024}\natexlab{}.
\newblock \showarticletitle{Toward re-identifying any animal}.
\newblock \bibinfo{journal}{\emph{Advances in Neural Information Processing Systems}}  \bibinfo{volume}{36} (\bibinfo{year}{2024}).
\newblock


\bibitem[Kern et~al\mbox{.}(2024)]%
        {kern2024towards}
\bibfield{author}{\bibinfo{person}{Daria Kern}, \bibinfo{person}{Tobias Schiele}, \bibinfo{person}{Ulrich Klauck}, {and} \bibinfo{person}{Winfred Ingabire}.} \bibinfo{year}{2024}\natexlab{}.
\newblock \showarticletitle{Towards Automated Chicken Monitoring: Dataset and Machine Learning Methods for Visual, Noninvasive Reidentification}.
\newblock \bibinfo{journal}{\emph{Animals}} \bibinfo{volume}{15}, \bibinfo{number}{1} (\bibinfo{year}{2024}), \bibinfo{pages}{1}.
\newblock


\bibitem[Leliveld et~al\mbox{.}(2022)]%
        {leliveld2022dairy}
\bibfield{author}{\bibinfo{person}{Lisette~MC Leliveld}, \bibinfo{person}{Elisabetta Riva}, \bibinfo{person}{Gabriele Mattachini}, \bibinfo{person}{Alberto Finzi}, \bibinfo{person}{Daniela Lovarelli}, {and} \bibinfo{person}{Giorgio Provolo}.} \bibinfo{year}{2022}\natexlab{}.
\newblock \showarticletitle{Dairy cow behavior is affected by period, time of day and housing}.
\newblock \bibinfo{journal}{\emph{Animals}} \bibinfo{volume}{12}, \bibinfo{number}{4} (\bibinfo{year}{2022}), \bibinfo{pages}{512}.
\newblock


\bibitem[Li et~al\mbox{.}(2018)]%
        {li2018richly}
\bibfield{author}{\bibinfo{person}{Dangwei Li}, \bibinfo{person}{Zhang Zhang}, \bibinfo{person}{Xiaotang Chen}, {and} \bibinfo{person}{Kaiqi Huang}.} \bibinfo{year}{2018}\natexlab{}.
\newblock \showarticletitle{A richly annotated pedestrian dataset for person retrieval in real surveillance scenarios}.
\newblock \bibinfo{journal}{\emph{IEEE Transactions on Image Processing}} \bibinfo{volume}{28}, \bibinfo{number}{4} (\bibinfo{year}{2018}), \bibinfo{pages}{1575--1590}.
\newblock


\bibitem[Li et~al\mbox{.}(2020)]%
        {li2020atrw}
\bibfield{author}{\bibinfo{person}{Shuyuan Li}, \bibinfo{person}{Jianguo Li}, \bibinfo{person}{Hanlin Tang}, \bibinfo{person}{Rui Qian}, {and} \bibinfo{person}{Weiyao Lin}.} \bibinfo{year}{2020}\natexlab{}.
\newblock \showarticletitle{ATRW: A Benchmark for Amur Tiger Re-identification in the Wild}. In \bibinfo{booktitle}{\emph{Proceedings of the 28th ACM International Conference on Multimedia}}. \bibinfo{pages}{2590--2598}.
\newblock


\bibitem[Li et~al\mbox{.}(2023)]%
        {li2023clip}
\bibfield{author}{\bibinfo{person}{Siyuan Li}, \bibinfo{person}{Li Sun}, {and} \bibinfo{person}{Qingli Li}.} \bibinfo{year}{2023}\natexlab{}.
\newblock \showarticletitle{CLIP-ReID: exploiting vision-language model for image re-identification without concrete text labels}. In \bibinfo{booktitle}{\emph{Proceedings of the AAAI Conference on Artificial Intelligence}}, Vol.~\bibinfo{volume}{37}. \bibinfo{pages}{1405--1413}.
\newblock


\bibitem[McVey et~al\mbox{.}(2023)]%
        {mcvey2023invited}
\bibfield{author}{\bibinfo{person}{Catherine McVey}, \bibinfo{person}{Fushing Hsieh}, \bibinfo{person}{Diego Manriquez}, \bibinfo{person}{Pablo Pinedo}, {and} \bibinfo{person}{Kristina Horback}.} \bibinfo{year}{2023}\natexlab{}.
\newblock \showarticletitle{Invited Review: Applications of unsupervised machine learning in livestock behavior: Case studies in recovering unanticipated behavioral patterns from precision livestock farming data streams}.
\newblock \bibinfo{journal}{\emph{Applied Animal Science}} \bibinfo{volume}{39}, \bibinfo{number}{2} (\bibinfo{year}{2023}), \bibinfo{pages}{99--116}.
\newblock


\bibitem[Papafitsoros et~al\mbox{.}(2021)]%
        {papafitsoros2021social}
\bibfield{author}{\bibinfo{person}{Kostas Papafitsoros}, \bibinfo{person}{Aliki Panagopoulou}, {and} \bibinfo{person}{Gail Schofield}.} \bibinfo{year}{2021}\natexlab{}.
\newblock \showarticletitle{Social media reveals consistently disproportionate tourism pressure on a threatened marine vertebrate}.
\newblock \bibinfo{journal}{\emph{Animal Conservation}} \bibinfo{volume}{24}, \bibinfo{number}{4} (\bibinfo{year}{2021}), \bibinfo{pages}{568--579}.
\newblock


\bibitem[Radford et~al\mbox{.}(2021)]%
        {radford2021learning}
\bibfield{author}{\bibinfo{person}{Alec Radford}, \bibinfo{person}{Jong~Wook Kim}, \bibinfo{person}{Chris Hallacy}, \bibinfo{person}{Aditya Ramesh}, \bibinfo{person}{Gabriel Goh}, \bibinfo{person}{Sandhini Agarwal}, \bibinfo{person}{Girish Sastry}, \bibinfo{person}{Amanda Askell}, \bibinfo{person}{Pamela Mishkin}, \bibinfo{person}{Jack Clark}, {et~al\mbox{.}}} \bibinfo{year}{2021}\natexlab{}.
\newblock \showarticletitle{Learning transferable visual models from natural language supervision}. In \bibinfo{booktitle}{\emph{International Conference on Machine Learning}}. PMLR, \bibinfo{pages}{8748--8763}.
\newblock


\bibitem[Schneider et~al\mbox{.}(2019)]%
        {schneider2019past}
\bibfield{author}{\bibinfo{person}{Stefan Schneider}, \bibinfo{person}{Graham~W Taylor}, \bibinfo{person}{Stefan Linquist}, {and} \bibinfo{person}{Stefan~C Kremer}.} \bibinfo{year}{2019}\natexlab{}.
\newblock \showarticletitle{Past, present and future approaches using computer vision for animal re-identification from camera trap data}.
\newblock \bibinfo{journal}{\emph{Methods in Ecology and Evolution}} \bibinfo{volume}{10}, \bibinfo{number}{4} (\bibinfo{year}{2019}), \bibinfo{pages}{461--470}.
\newblock


\bibitem[Schofield et~al\mbox{.}(2022)]%
        {schofield2022more}
\bibfield{author}{\bibinfo{person}{Gail Schofield}, \bibinfo{person}{Kostas Papafitsoros}, \bibinfo{person}{Chloe Chapman}, \bibinfo{person}{Akanksha Shah}, \bibinfo{person}{Lucy Westover}, \bibinfo{person}{Liam~CD Dickson}, {and} \bibinfo{person}{Kostas~A Katselidis}.} \bibinfo{year}{2022}\natexlab{}.
\newblock \showarticletitle{More aggressive sea turtles win fights over foraging resources independent of body size and years of presence}.
\newblock \bibinfo{journal}{\emph{Animal Behaviour}}  \bibinfo{volume}{190} (\bibinfo{year}{2022}), \bibinfo{pages}{209--219}.
\newblock


\bibitem[Shi et~al\mbox{.}(2022)]%
        {shi2022dense}
\bibfield{author}{\bibinfo{person}{Xinyu Shi}, \bibinfo{person}{Dong Wei}, \bibinfo{person}{Yu Zhang}, \bibinfo{person}{Donghuan Lu}, \bibinfo{person}{Munan Ning}, \bibinfo{person}{Jiashun Chen}, \bibinfo{person}{Kai Ma}, {and} \bibinfo{person}{Yefeng Zheng}.} \bibinfo{year}{2022}\natexlab{}.
\newblock \showarticletitle{Dense cross-query-and-support attention weighted mask aggregation for few-shot segmentation}. In \bibinfo{booktitle}{\emph{European Conference on Computer Vision}}. Springer, \bibinfo{pages}{151--168}.
\newblock


\bibitem[Van~der Maaten and Hinton(2008)]%
        {van2008visualizing}
\bibfield{author}{\bibinfo{person}{Laurens Van~der Maaten} {and} \bibinfo{person}{Geoffrey Hinton}.} \bibinfo{year}{2008}\natexlab{}.
\newblock \showarticletitle{Visualizing data using t-SNE.}
\newblock \bibinfo{journal}{\emph{Journal of Machine Learning Research}} \bibinfo{volume}{9}, \bibinfo{number}{11} (\bibinfo{year}{2008}).
\newblock


\bibitem[Varghese et~al\mbox{.}(2023)]%
        {varghese2023fine}
\bibfield{author}{\bibinfo{person}{Anjli Varghese}, \bibinfo{person}{Malathy Jawahar}, {and} \bibinfo{person}{A~Amalin Prince}.} \bibinfo{year}{2023}\natexlab{}.
\newblock \showarticletitle{Fine-tuning ConvNets with novel leather image data for species identification}. In \bibinfo{booktitle}{\emph{Fifteenth International Conference on Machine Vision}}, Vol.~\bibinfo{volume}{12701}. SPIE, \bibinfo{pages}{150--157}.
\newblock


\bibitem[Wahltinez and Wahltinez(2024)]%
        {wahltinez2024open}
\bibfield{author}{\bibinfo{person}{Oscar Wahltinez} {and} \bibinfo{person}{Sarah~J Wahltinez}.} \bibinfo{year}{2024}\natexlab{}.
\newblock \showarticletitle{An open-source general purpose machine learning framework for individual animal re-identification using few-shot learning}.
\newblock \bibinfo{journal}{\emph{Methods in Ecology and Evolution}} \bibinfo{volume}{15}, \bibinfo{number}{2} (\bibinfo{year}{2024}), \bibinfo{pages}{373--387}.
\newblock


\bibitem[Wang and Liu(2021)]%
        {wang2021understanding}
\bibfield{author}{\bibinfo{person}{Feng Wang} {and} \bibinfo{person}{Huaping Liu}.} \bibinfo{year}{2021}\natexlab{}.
\newblock \showarticletitle{Understanding the behaviour of contrastive loss}. In \bibinfo{booktitle}{\emph{Proceedings of the IEEE/CVF Conference on Computer Vision and Pattern Recognition}}. \bibinfo{pages}{2495--2504}.
\newblock


\bibitem[Weideman et~al\mbox{.}(2020)]%
        {weideman2020extracting}
\bibfield{author}{\bibinfo{person}{Hendrik Weideman}, \bibinfo{person}{Chuck Stewart}, \bibinfo{person}{Jason Parham}, \bibinfo{person}{Jason Holmberg}, \bibinfo{person}{Kiirsten Flynn}, \bibinfo{person}{John Calambokidis}, \bibinfo{person}{D~Barry Paul}, \bibinfo{person}{Anka Bedetti}, \bibinfo{person}{Michelle Henley}, \bibinfo{person}{Frank Pope}, {et~al\mbox{.}}} \bibinfo{year}{2020}\natexlab{}.
\newblock \showarticletitle{Extracting identifying contours for African elephants and humpback whales using a learned appearance model}. In \bibinfo{booktitle}{\emph{Proceedings of the IEEE/CVF Winter Conference on Applications of Computer Vision}}. \bibinfo{pages}{1276--1285}.
\newblock


\bibitem[Xu et~al\mbox{.}(2024)]%
        {xu2024advanced}
\bibfield{author}{\bibinfo{person}{Pengfei Xu}, \bibinfo{person}{Yuanyuan Zhang}, \bibinfo{person}{Minghao Ji}, \bibinfo{person}{Songtao Guo}, \bibinfo{person}{Zhanyong Tang}, \bibinfo{person}{Xiang Wang}, \bibinfo{person}{Jing Guo}, \bibinfo{person}{Junjie Zhang}, {and} \bibinfo{person}{Ziyu Guan}.} \bibinfo{year}{2024}\natexlab{}.
\newblock \showarticletitle{Advanced intelligent monitoring technologies for animals: A survey}.
\newblock \bibinfo{journal}{\emph{Neurocomputing}}  \bibinfo{volume}{585} (\bibinfo{year}{2024}), \bibinfo{pages}{127640}.
\newblock


\bibitem[Yu et~al\mbox{.}(2024)]%
        {yu2024rethinking}
\bibfield{author}{\bibinfo{person}{Han Yu}, \bibinfo{person}{Xingxuan Zhang}, \bibinfo{person}{Renzhe Xu}, \bibinfo{person}{Jiashuo Liu}, \bibinfo{person}{Yue He}, {and} \bibinfo{person}{Peng Cui}.} \bibinfo{year}{2024}\natexlab{}.
\newblock \showarticletitle{Rethinking the evaluation protocol of domain generalization}. In \bibinfo{booktitle}{\emph{Proceedings of the IEEE/CVF Conference on Computer Vision and Pattern Recognition}}. \bibinfo{pages}{21897--21908}.
\newblock


\bibitem[Zhang et~al\mbox{.}(2024)]%
        {zhang2024visual}
\bibfield{author}{\bibinfo{person}{Jialu Zhang}, \bibinfo{person}{Xinyi Wang}, \bibinfo{person}{Chenglin Yao}, \bibinfo{person}{Jianfeng Ren}, {and} \bibinfo{person}{Xudong Jiang}.} \bibinfo{year}{2024}\natexlab{}.
\newblock \showarticletitle{Visual-linguistic Cross-domain Feature Learning with Group Attention and Gamma-correct Gated Fusion for Extracting Commonsense Knowledge}. In \bibinfo{booktitle}{\emph{Proceedings of the 32nd ACM International Conference on Multimedia}}. \bibinfo{pages}{4650--4659}.
\newblock


\bibitem[Zhang et~al\mbox{.}(2021)]%
        {zhang2021yakreid}
\bibfield{author}{\bibinfo{person}{Tingting Zhang}, \bibinfo{person}{Qijun Zhao}, \bibinfo{person}{Cuo Da}, \bibinfo{person}{Liyuan Zhou}, \bibinfo{person}{Lei Li}, {and} \bibinfo{person}{Suonan Jiancuo}.} \bibinfo{year}{2021}\natexlab{}.
\newblock \showarticletitle{Yakreid-103: A benchmark for yak re-identification}. In \bibinfo{booktitle}{\emph{2021 IEEE International Joint Conference on Biometrics}}. IEEE, \bibinfo{pages}{1--8}.
\newblock


\bibitem[Zhao et~al\mbox{.}(2024)]%
        {zhao2024symmetric}
\bibfield{author}{\bibinfo{person}{Di Zhao}, \bibinfo{person}{Yun~Sing Koh}, \bibinfo{person}{Gillian Dobbie}, \bibinfo{person}{Hongsheng Hu}, {and} \bibinfo{person}{Philippe Fournier-Viger}.} \bibinfo{year}{2024}\natexlab{}.
\newblock \showarticletitle{Symmetric Self-Paced Learning for Domain Generalization}. In \bibinfo{booktitle}{\emph{Proceedings of the AAAI Conference on Artificial Intelligence}}, Vol.~\bibinfo{volume}{38}. \bibinfo{pages}{16961--16969}.
\newblock


\end{thebibliography}

\end{document}